\documentclass[letterpaper, 10 pt, conference]{ieeeconf}
\DeclareUnicodeCharacter{0301}{\'{e}}
\IEEEoverridecommandlockouts
\usepackage{cite}
\usepackage{amsmath,amssymb,amsfonts}
\usepackage{algorithmic}
\usepackage{graphicx}
\usepackage{textcomp}
\usepackage{xcolor}
\usepackage{tabularx}
\usepackage{booktabs,multirow}
\usepackage{adjustbox}
\usepackage[section]{placeins}
\usepackage{dblfloatfix}
\usepackage{float}
\usepackage{arydshln}
\usepackage{url}
\usepackage{hyperref}
\def\BibTeX{{\rm B\kern-.05em{\sc i\kern-.025em b}\kern-.08em
    T\kern-.1667em\lower.7ex\hbox{E}\kern-.125emX}}

\newcommand{\expone}{Indoor (Dark+HDR)}
\newcommand{\exptwo}{Indoor (Dark+ Blur)}
\newcommand{\expthree}{Indoor (HDR + Blur)}
\newcommand{\expfour}{Outdoor (Well Lit + Blur)}
\newcommand{\expfive}{Outdoor (Dark + Blur)}
\newcommand{\expsix}{Indoor (Drone Flying)}
\newcommand{\expseven}{Motorcycle (Fast Motion)}
\newcommand{\expeight}{Outdoor (Low Brightness)}
\newcommand{\expnine}{Outdoor (Dark)}
\newcommand{\nameframew}{EECVS }

\begin{document}

\title{Efficient Event Camera Volume System}

\author{
Juan Soto$^{1*}$, Ian Noronha$^{1*}$, Saru Bharti$^{1}$, Upinder Kaur$^{1\dagger}$
\thanks{$^{1}$Purdue University, USA. \texttt{\{inoronha, soto97, bharti3, kauru\}@purdue.edu}}
\thanks{$^{*}$These authors contributed equally to this work. $\dagger$ Corresponding author.}
\thanks{Project page and code: \url{https://github.com/Dookiep/EECVS.git}.}
 }

\maketitle
\begin{abstract}
Event cameras promise low latency and high dynamic range, yet their sparse output challenges integration into standard robotic pipelines. We introduce \nameframew (Efficient Event Camera Volume System), a novel framework that models event streams as continuous-time Dirac impulse trains, enabling artifact-free compression through direct transform evaluation at event timestamps. Our key innovation combines density-driven adaptive selection among DCT, DTFT, and DWT transforms with transform-specific coefficient pruning strategies tailored to each domain's sparsity characteristics. The framework eliminates temporal binning artifacts while automatically adapting compression strategies based on real-time event density analysis. On EHPT-XC and MVSEC datasets, our framework achieves superior reconstruction fidelity with DTFT delivering the lowest earth mover distance. In downstream segmentation tasks, EECVS demonstrates robust generalization. Notably, our approach demonstrates exceptional cross-dataset generalization: when evaluated with EventSAM segmentation, EECVS achieves mean IoU 0.87 on MVSEC versus 0.44 for voxel grids at 24 channels, while remaining competitive on EHPT-XC. Our ROS2 implementation provides real-time deployment with DCT processing achieving 1.5 ms latency and 2.7× higher throughput than alternative transforms, establishing the first adaptive event compression framework that maintains both computational efficiency and superior generalization across diverse robotic scenarios.

\end{abstract}


\section{Introduction}

Robotic systems are increasingly required to operate in environments that place strong demands on perception, such as high-speed navigation, low-light conditions, or scenes with extreme contrast. Standard RGB cameras suffer fundamental limitations in these scenarios: motion blur corrupts fast movements, limited dynamic range loses critical information, and high processing overheads constrain real-time performance \bstctlcite{BSTcontrol}\cite{wen_autonomous_2019} \cite{xiao_making_2020}. These constraints fundamentally limit robotic perception reliability and restrict autonomous operation in dynamic environments. 

Event cameras address these limitations through asynchronous, pixel-independent brightness change detection. Unlike traditional cameras capturing full frames at fixed intervals, event cameras generate continuous streams of spatiotemporal events containing coordinates, timestamps, and polarity information. This sensing paradigm enables key properties, such as microsecond temporal resolution, high dynamic range of up to 140 dB, and low power consumption~\cite{gallego_event-based_2022}. These characteristics make event cameras particularly suitable for robotic applications that require fast and reliable perception in visually demanding scenarios with limited computational resources~\cite{gallego_event-based_2022}.



\begin{figure}[!t]  
  \centering
  \includegraphics[width=1\linewidth]{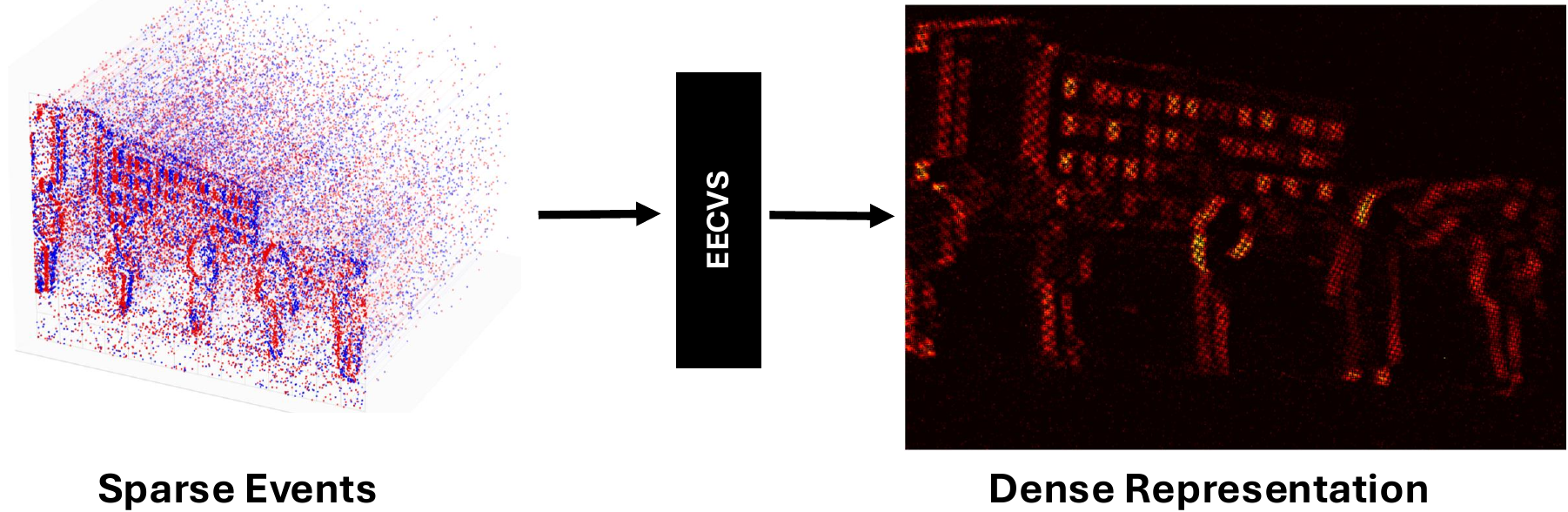}
  \caption{\textbf{Event-to-dense representation in EECVS.} Incoming event streams are processed within the framework and converted into compact dense representations through the application of DCT, DTFT, or DWT.}\vspace{-0.5em}
  \label{fig:framework-arch}
  \end{figure}

Despite these advantages, the sparse and asynchronous nature of event data creates a fundamental integration challenge: standard vision pipelines expect dense inputs, while events arrive as irregular spatiotemporal streams. Current approaches employ fixed representations that ignore stream variability. Others convert sparse events into dense images for efficient processing, but sacrifice fine temporal resolution~\cite{gehrig_recurrent_2023}. Voxelized representations preserve spatiotemporal structure yet blur information through temporal binning. TORE volumes encode only recent events, discarding longer temporal dependencies~\cite{baldwin_time-ordered_2023}. CES volumes apply single transforms, such as DFT (Discrete Fourier Transform), for compact encodings, but lack the flexibility to adapt across diverse scenarios~\cite{lin_compressed_2025}. These highlight the critical need for adaptive compression frameworks that can intelligently respond to varying event stream characteristics.


\begin{figure*}[!t]  
  \centering
  \includegraphics[width=1\linewidth]{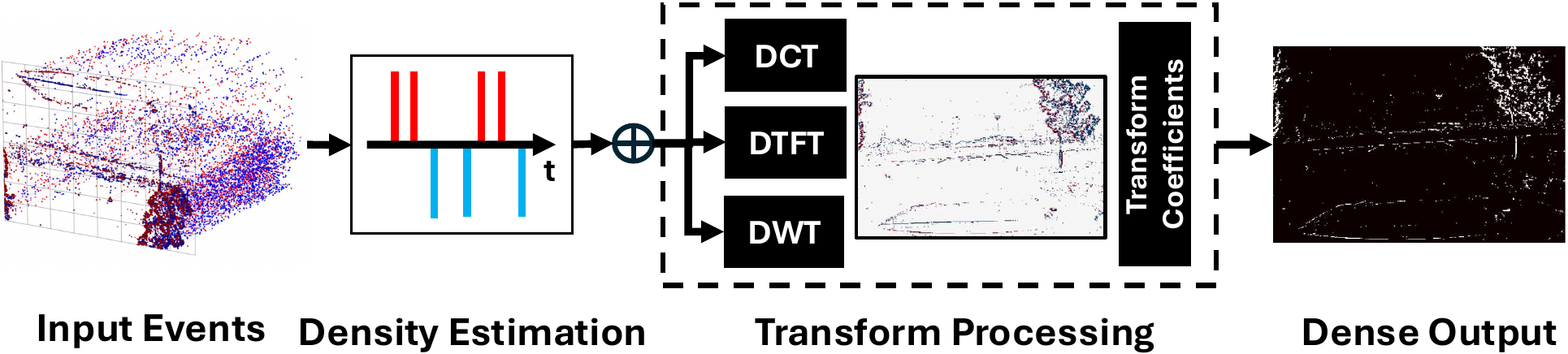}
  \caption{Compression process for a single event window. Events are aggregated, transformed with a basis selected according to activity density, pruned by either low-frequency retention (DCT) or magnitude selection (DTFT/DWT), and packed into dense representations. }
  \label{fig:method-fig}
\end{figure*}

We address this challenge through \nameframew, a unified framework that adaptively compresses event streams using autonomous density-driven transform selection. Our framework leverages compressed sensing theory, which proves that sparse signals in appropriate domains require few coefficients for accurate reconstruction. \nameframew exploits event stream sparsity through intelligent, density-aware transform selection for real-time adaptation to changing scene dynamics. As illustrated in Fig.~\ref{fig:framework-arch}, the framework ingests event streams and outputs dense representations that serve as the basis for robotic perception. The framework adaptively chooses among three complementary transforms: the \textbf{Discrete Cosine Transform (DCT)} excels at energy concentration in high-density regions, the \textbf{Discrete Time Fourier Transform (DTFT)} preserves temporal fidelity for moderate event rates, and the \textbf{Discrete Wavelet Transform (DWT)} maintains localized structure for sparse activity patterns. Our approach models event streams as continuous-time Dirac impulse trains, enabling direct transform evaluation at event timestamps without temporal binning artifacts that compromise reconstruction quality. This mathematical foundation, combined with density-driven selection criteria grounded in information-theoretic principles, produces dense representations that minimize information loss while ensuring compatibility with robotic perception pipelines.


In this paper, we address the need for efficient and adaptable event representations in robotics and present a unified compression framework. Our contributions are:
\begin{itemize}
\item \textbf{EECVS autonomous compression framework:} An intelligent framework that combines continuous-time Dirac impulse modeling with density-driven transform selection, eliminating temporal binning artifacts while enabling task-adaptive performance across diverse robotic scenarios with computational efficiency.  
\item \textbf{Systematic Reconstruction Analysis: } Comprehensive evaluation demonstrating how each transform preserves temporal fidelity, spatial detail, and information content under varying event densities, providing insights for optimal compression strategy selection. 
\item \textbf{Downstream Validation and Real-world Deployment:} Extensive downstream task evaluation revealing superior generalization capabilities, coupled with a ROS2 implementation supporting real-world robotic deployment. 
\end{itemize}

\section{Related Work}


Early event processing methods faced fundamental trade-offs between temporal precision and computational tractability. Sequential processing approaches achieve microsecond latency through probabilistic filters and spiking neural models~\cite{lagorce_hots_2017, orchard_converting_2015}. These techniques update state sequentially using continuous time dynamics, but their reliance on handcrafted rules and parameter tuning limits their scalability to higher-level perception tasks. Spiking Neural Networks (SNN) extend this paradigm with data-driven approaches ~\cite{zhao_feedforward_2015,zhu_event-based_2022}, but practical use remains difficult due to training instabilities. 

Aggregation-based methods trade temporal precision for computational efficiency and compatibility with standard vision pipelines. Time Ordered Recent Event (TORE) \cite{baldwin_time-ordered_2023} representations store short temporal histories in timestamp grids to reduce update costs and capture local temporal sequences. However, it still quantizes past information and loses fine detail, thereby eroding the benefits of using event cameras. Other approaches aggregate events into grid or voxel-based structures that accumulate event counts or polarities over fixed intervals, producing dense representations that are compatible with convolutional networks~\cite{zhu_ev-flownet_2018,rebecq_events--video_2019,gehrig_end--end_2019}. These methods intuitively highlight brightness changes and edges but inevitably discard fine timing by merging events into coarse temporal bins. 

Recent research explores improved event representations and learning-based processing strategies. ERGO-12~\cite{zubic_chaos_2023} studies how the choice of dense event encodings affects downstream learning performance and proposes selecting representations using the Gromov–Wasserstein discrepancy. Other approaches address the variability of event density through adaptive stacking strategies, such as multi-density event stacks (MDES)~\cite{nam_stereo_2022}. Event focal stack representations~\cite{lou_all--focus_2023} further exploit the temporal continuity of events to encode scene structure across time. In parallel, other works represent events as sets of spatiotemporal points processed by graph networks or transformers \cite{sabater_event_2022,schaefer_aegnn_2022,xie_event_2024}, preserving sparsity and temporal resolution. However, these approaches increase computational overheads and are often tailored toward specific perception tasks.


Compressed sensing provides the theoretical foundation for our framework, adapting compression strategies across multiple domains. Medical imaging pioneered adaptive compression through Sparse MRI, which dynamically adjusts acquisition strategies \cite{lustig_sparse_2007}. Computed tomography employs compressed sensing for adaptive low-dose reconstruction \cite{konovalov_compressed-sensing-inspired_2024}. Seismic imaging in geophysics applies scene-dependent processing for subsurface recovery \cite{herrmann_fighting_2012}. These successes demonstrate that adaptive selection among transforms enhances compression effectiveness when matched to signal characteristics. 

Event data exhibits natural spatiotemporal sparsity, making it ideal for adaptive compression frameworks. However, existing approaches lack principled selection criteria for choosing among transform bases. Our framework addresses this gap by establishing density-driven selection criteria grounded in compressed sensing theory, enabling adaptive compression that responds to varying event stream characteristics.




\section{Methods}

\nameframew adaptively compresses event streams through density-driven transform selection and coefficient pruning strategies tailored to each transform's characteristics. The framework continuously monitors activity levels within temporal windows. This real-time analysis drives autonomous selection among three specialized transforms. We derive an event density measure that helps classify the perceived event streams. For sparse windows, the framework deploys wavelets to preserve isolated features. When detecting moderate activity, \nameframew switches to Fourier analysis for temporal fidelity. Dense activity triggers the framework's cosine transform for maximum energy compaction. Following transform selection, \nameframew autonomously reduces coefficients to a proportion $r=\min|M,\mathcal{K}_w|$, using transform-specific pruning strategies. For DCT, \nameframew retains the first $r$ low-frequency indices. For DTFT and DWT, the system selects the $r$ largest-magnitude coefficients. The framework then packages these compact descriptors into dense representations compatible with standard perception pipelines (Fig.~\ref{fig:method-fig}).
  
\subsection{Event Density Estimation and Transform Selection}\vspace{-0.25em}

To adaptively compress event streams, we introduce an event density measure that quantifies the activity levels within temporal windows and provides the basis for principled transform selection. Let the input window be
\[
E_w = \{(t_i, x_i, y_i, p_i)\}_{i=1}^{N_w},
\]
where $N_w = |E_w|$ is the number of events observed in window $w$, $t_i$ is the timestamp, $(x_i, y_i)$ are pixel coordinates and $p_i \in \{-1,+1\}$ is the polarity. For a sensor of height $H$ and width $W$, and a window duration $T$, the normalized event density is defined as
\begin{equation}
    \rho_w = \frac{|E_w|}{H \cdot W \cdot T},
    \label{eq:event_density}
\end{equation}
expressed events per pixel per unit time (events/s). This normalization allows consistent comparisons across different sensor resolutions and temporal scales.

We establish three density regimes based on empirical analysis of event distribution characteristics. Our threshold selection uses percentile-based partitioning, which adapts naturally to sensor and scene characteristics while providing stable selection criteria. We partition event streams according to thresholds $\tau_{\text{low}}$ and $\tau_{\text{high}}$:
\begin{equation}
\text{Regime}(w) =
\begin{cases}
\text{Sparse}, & \rho_w < \tau_{\text{low}},\\
\text{Moderate}, & \tau_{\text{low}} \le \rho_w < \tau_{\text{high}},\\
\text{Dense}, & \rho_w \ge \tau_{\text{high}}.
\end{cases}
\label{eq:regimes}
\end{equation}

Thresholds $\tau_{\text{low}}$ and $\tau_{\text{high}}$ are set on a calibration split using percentiles of the window density $\rho_w$ in \eqref{eq:event_density}. We compute the empirical distribution of $\rho_w$ over all windows in the split and define $\tau_{\text{low}}$ as the 25th percentile and $\tau_{\text{high}}$ as the 75th percentile. This choice ensures balanced workload distribution while maintaining theoretical grounding:  the lowest quartile typically represents sparse, isolated events best suited for localized wavelet representation; the highest quartile contains dense activity requiring energy concentration through cosine transforms; the middle half exhibits moderate structure optimally preserved by Fourier analysis.  With this choice, windows in the lowest quartile are assigned to DWT, windows in the highest quartile are assigned to DCT, and the middle half is assigned to DTFT, consistent with \eqref{eq:regimes} and \eqref{eq:transform_select}. Percentile thresholds adapt to the sensor and scene without manual tuning, since $\rho_w$ is normalized by $H$, $W$, and $T$. After calibration, thresholds remain fixed during deployment to ensure consistent selection behavior.

For each density regime, the framework selects the transform domain that yields the sparsest and most informative representation:
\begin{equation}
\text{Transform}(w) =
\begin{cases}
\mathrm{DWT}, & \rho_w < \tau_{\text{low}},\\
\mathrm{DTFT}, & \tau_{\text{low}} \le \rho_w < \tau_{\text{high}},\\
\mathrm{DCT}, & \rho_w \ge \tau_{\text{high}}.
\end{cases}
\label{eq:transform_select}
\end{equation}

This selection strategy exploits the complementary sparsity of DWT, DTFT, and DCT across density regimes, forming the basis for event-driven transform encoding.

\subsection{Event-Driven Signal Model and Transform Encoding}
\begin{figure}
    \centering
    \includegraphics[width=1\linewidth]{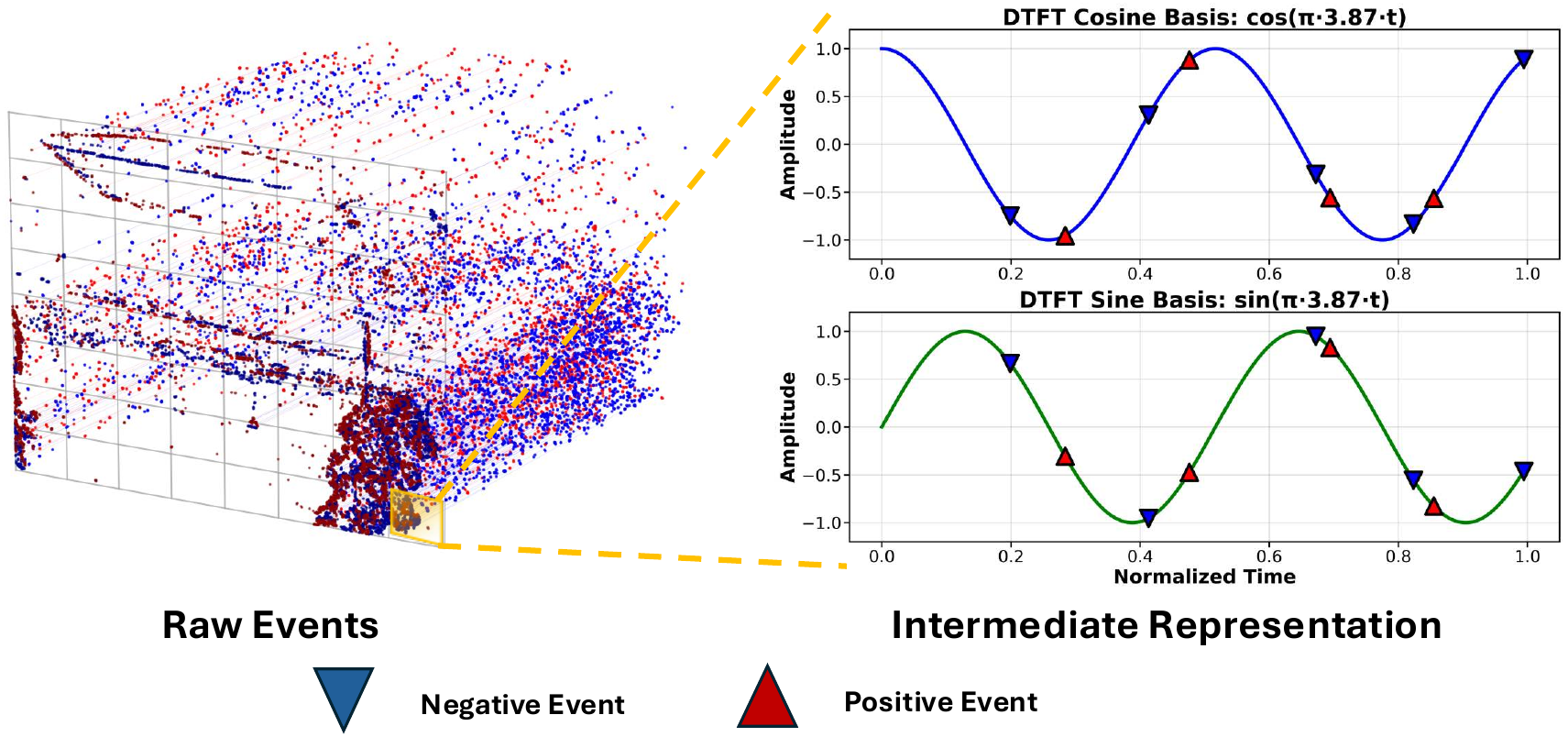}
    \caption{Per pixel DTFT within a window. Events sample transform bases directly, and coefficients are pruned according to the transform-specific retention strategy.}
    \label{fig:pixel-ecoding}
\end{figure}

\label{sec:event-driven-encoding}

\paragraph{Dirac impulse model.}
Within a window \(w\), we model the event stream as a sum of Dirac impulses in continuous time, avoiding temporal binning artifacts:
\begin{equation}
s_w(t) \;=\; \sum_{i=1}^{N_w} p_i\,\delta\!\left(t - t_i\right),
\label{eq:dirac_model}
\end{equation}

where each event $(t_i, x_i, y_i, p_i)$ contributes polarity $p_i\in\{-1,+1\}$ at timestamp $t_i$. This formulation enables direct evaluation of transform atoms at event timestamps, preserving temporal precision while maintaining computational efficiency  (see Fig.~\ref{fig:pixel-ecoding}). EECVS processes each pixel independently: for every spatial location $(x,y)$ we accumulate basis samples from that pixel's events within the current window. The transform choice derives from window-level density analysis, while coefficient computation occurs at the pixel level.

\paragraph{Coefficient accumulation by basis sampling.}
Let \(\{\phi_k(t)\}_{k\in\mathcal{K}_w}\) denote the transform atoms selected for window \(w\) by the density-driven selection rule (Section above, cf.\ Eq.~\eqref{eq:event_density}). Using the distributional property of the Dirac delta function \(\delta\), we compute coefficients as:
\begin{equation}
\begin{aligned}
c_{w,k} &= \langle s_w,\phi_k\rangle \\
        &= \int s_w(t)\,\phi_k(t)\,dt \\
        &= \sum_{i=1}^{N_w} p_i\,\phi_k(t_i),
        \quad k\in\mathcal{K}_w .
\end{aligned}
\label{eq:coeff_accum}
\end{equation}

We order \(\mathcal{K}_w\) to prioritize low frequencies or coarse scales, which concentrate energy and promote sparsity in the compressed representation.

\paragraph{Transforms by regime.}
Given \(\mathrm{Transform}(w)\in\{\mathrm{DWT},\mathrm{DTFT},\mathrm{DCT}\}\), each transform provides distinct advantages:
\begin{itemize}
  \item \textbf{Sparse regime (DWT).} Wavelet atoms \(\psi_{j,m}(t)\) at coarse scales \(j\) and shifts \(m\) provide localized representation ideal for isolated events:
  \(
  c_{w,(j,m)}=\sum_i p_i\,\psi_{j,m}(t_i)
  \),
  with \((j,m)\in\mathcal{K}_w\) biased toward coarser scales to preserve isolated activity structure.
  \item \textbf{Moderate regime (DTFT).} Complex sinusoids \(\phi_\omega(t)=e^{-\,\mathrm{i}\omega t}\) over a finite set of low angular frequencies \(\omega\in\Omega_w\) maintain temporal fidelity:
  \(
  c_{w,\omega}=\sum_i p_i\,e^{-\,\mathrm{i}\omega t_i}
  \).
  We retain the lowest \(|\omega|\) values to preserve temporal structure with compact spectral representation.
  \item \textbf{Dense regime (DCT).} Real cosine atoms \(\phi_k(t)=\cos(\omega_k t)\) with ordered frequencies \(\omega_k\)  provide efficient energy compaction:
  \(
  c_{w,k}=\sum_i p_i \cos(\omega_k t_i)
  \),
  keeping the lowest indices \(k\in\mathcal{K}_w\) for strong energy concentration in dense activity patterns.
\end{itemize}

\subsection{Coefficient Retention and Representation Packing}


We use a fixed coefficient budget of $M$ per window. Unless stated otherwise, we use $M=16$, as it provides a balance between compression efficiency and reconstruction quality. For each window, we retain
\[
r=\min\{M,\,|\mathcal{K}_w|\},
\]
ensuring that the number of selected coefficients does not exceed the available basis size. With this budget established, coefficient pruning strategies match transform characteristics:

For \textbf{DCT}, we use frequency retention. We order cosine indices from low to high and keep the first $r$. This promotes energy compaction in dense activity. 
For \textbf{DTFT} and \textbf{DWT}, we conduct magnitude selection. We rank all coefficients by $|c_{w,k}|$ over the chosen frequency grid (DTFT) or over the scale and shift atoms (DWT) and keep the top $r$. This is robust in sparse or moderately populated windows where energy is not localized at low frequency. Ties are resolved in favor of the lower frequency or coarser scale. 

\begin{table}[t]
\centering
\small
\setlength{\tabcolsep}{3pt}
\renewcommand{\arraystretch}{0.92}
\caption{Overall means for two coefficient sizes using spatiotemporal metrics (MSE↓, SSIM↑, EMD↓).}\vspace{-0.25em}
\label{tab:overall-means-M}
\begin{adjustbox}{max width=\columnwidth} 
\begin{tabular}{lcccc}
\toprule
Method & $M$ & MSE ($\times10^{-3}$) & SSIM & EMD \\
\midrule
\multirow{2}{*}{DCT}  & 8  & \textbf{1.731} & \textbf{0.850} & 0.964 \\
                      & 24 & 1.740           & 0.849          & 0.964 \\
\multirow{2}{*}{DTFT} & 8  & 1.389           & 0.882          & 0.964 \\
                      & 24 & \textbf{1.350}  & \textbf{0.887} & 0.964 \\
\multirow{2}{*}{DWT}  & 8  & \textbf{2.010}  & 0.819          & 0.964 \\
                      & 24 & 2.011           & 0.819          & 0.964 \\
\bottomrule
\end{tabular}
\end{adjustbox}
\end{table}

The packed descriptor is $\hat{\mathbf{c}}_w=(c_{w,k})_{k\in \mathcal{K}_w^\alpha}$, with $\mathcal{K}_w^\alpha$ equal to the low frequency set for DCT and the top magnitude set for DTFT or DWT. 
We also report ablations with $M=8$ and $M=24$ to study the trade off between bitrate and fidelity. 
Windows are processed independently and then concatenated for the downstream perception stack. The effect of $M$ on reconstruction is summarized in Table~\ref{tab:overall-means-M}.

\subsection{ROS2 implementation}

We implemented EECVS as a modular ROS2 framework for seamless real-time robotic deployment. The system comprises three autonomous components operating together to provide comprehensive event stream processing. First, an event emulator generates configurable synthetic streams, enabling reproducible testing without specialized hardware. Second, a density estimation module continuously monitors event distributions and publishes real-time metrics that drive transform selection. Third, a logging module records compression features and coefficients, enabling offline analysis of the framework's autonomous decisions.

The implementation follows ROS2 conventions to ensure integration compatibility with existing robotic perception stacks. EECVS's modular architecture supports substitution of compression operators. Developers can extend the framework with additional transforms while preserving autonomous selection. This design enables flexible deployment across simulation and physical robots. 

\section{Experiments and Results}

\subsection{Experimental Setup}

We evaluate EECVS against established baselines to demonstrate autonomous compression effectiveness across diverse scenarios. Our framework competes with CES volumes and TORE volumes, representing current state-of-the-art approaches to event compression and representation. For EECVS evaluation, we configure $M=16$ retained coefficients.

\begin{figure}[t]
    \centering
    \includegraphics[width=\linewidth]{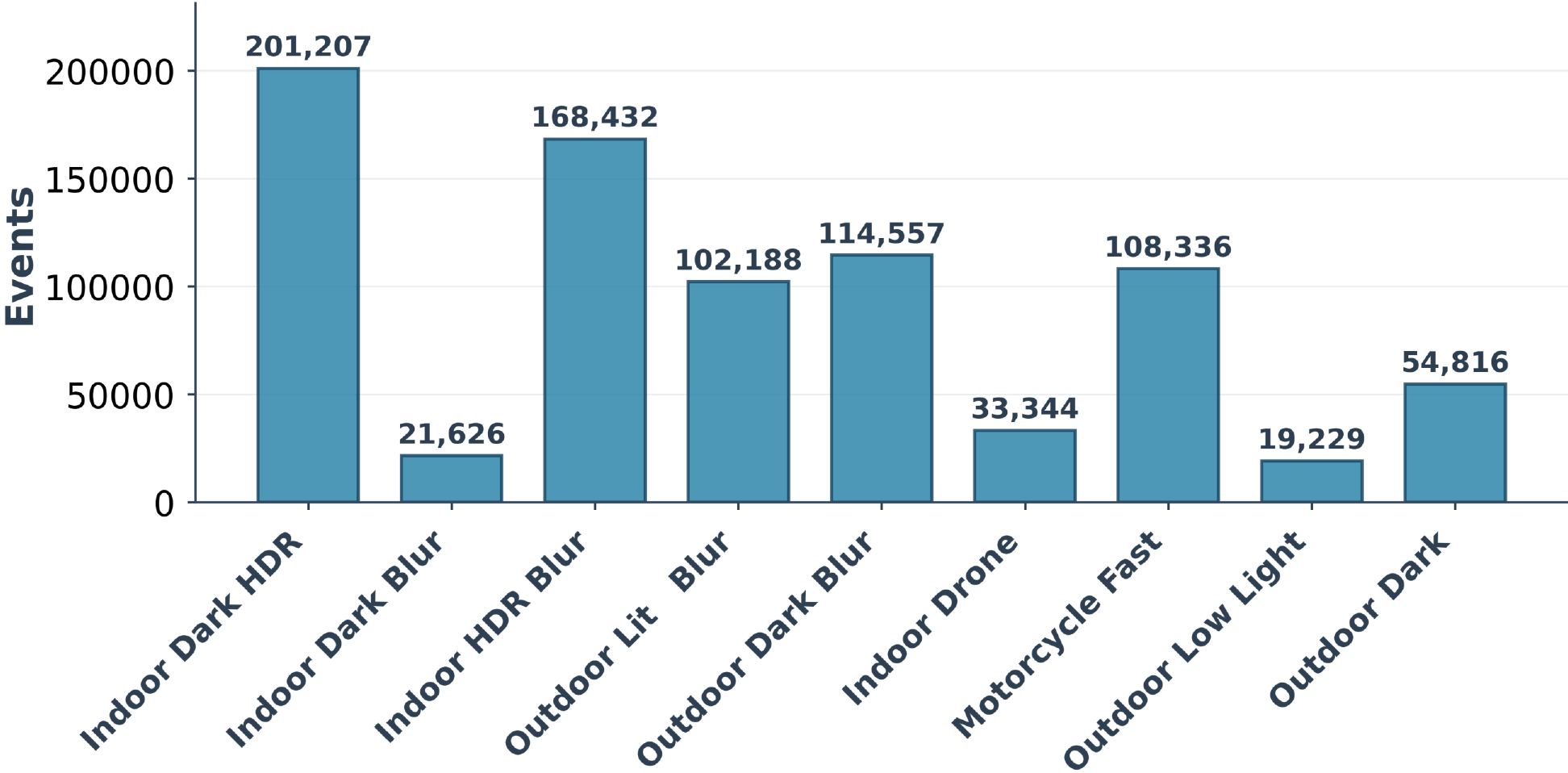}
    \caption{Histogram of event densities across the EHPT-XC and MVSEC datasets. 
    EHPT-XC exhibits high-density real-world sequences, while MVSEC contains more moderate motion-driven densities.}
    \label{fig:event_density_hist}
\end{figure}

Experiments span EHPT-XC \cite{cho_benchmark_2024} and MVSEC \cite{zhu_multi_2018} datasets. These datasets challenge the framework with significantly different density characteristics (Fig.~\ref{fig:event_density_hist}), with EHPT-XC containing high-density real-world sequences and MVSEC providing moderate motion-driven densities. Nine representative scenarios test the framework's autonomous adaptation across indoor/outdoor environments, varying lighting conditions, and different motion patterns. This evaluation design reveals how EECVS balances compression efficiency, reconstruction quality, and computational robustness.

\subsection{Reconstruction Performance Analysis}\label{subsec:quality-reconstruction}

Reconstruction metrics quantify EECVS's information preservation capabilities through adaptive selection of compression schemes. The framework reconstructs signals by inverse-transforming retained coefficients, enabling direct comparison with original event streams. We measure Mean Squared Error (MSE) and Structural Similarity Index (SSIM) on rendered frames, while Earth Mover Distance (EMD) evaluates temporal alignment preservation.

Table~\ref{tab:cross_metrics_wide_tech_cols} demonstrates that EECVS's DTFT selection delivers superior performance across diverse scenarios. The framework achieves the smallest EMD in eight of nine experiments while maintaining the lowest MSE and highest SSIM values. In Indoor (Dark+HDR), DTFT achieves remarkable reconstruction fidelity with MSE 0.0001, SSIM 0.987, and EMD 1.102. For Indoor (HDR+Blur), EECVS again delivers exceptional performance with MSE 0.0001 and EMD 0.642, significantly outperforming alternative approaches.

The framework demonstrates intelligent adaptation to sparse conditions through transform switching. In Outdoor (Low Brightness), DWT selection improves temporal alignment with EMD 1.063 versus 1.078 for DTFT, showcasing EECVS's ability to optimize for specific scene characteristics. CES occasionally achieves lower MSE in dynamic scenes but sacrifices temporal fidelity, as evidenced by substantially higher EMD values. TORE exhibits consistently higher errors across all metrics, confirming EECVS's superior compression effectiveness.

These results validate the framework's density-driven adaptive selection strategy: EECVS defaults to DTFT for balanced accuracy across most scenarios, automatically switches to DWT when detecting sparse event patterns, and triggers DCT mode for maximum compression efficiency in dense conditions.

\begin{table*}[!t]
\centering
\small 
\vspace{3mm}
\caption{Comparison of compression techniques using spatiotemporal metrics: spatial (MSE↓, SSIM↑) and temporal (EMD↓).}

\label{tab:cross_metrics_wide_tech_cols}
\begin{adjustbox}{max width=\textwidth}
\begin{tabular}{@{}l *{5}{r r r}@{}}
\toprule
& \multicolumn{3}{c}{\textbf{DCT}} & \multicolumn{3}{c}{\textbf{DTFT}}
& \multicolumn{3}{c}{\textbf{DWT}} & \multicolumn{3}{c}{\textbf{CES}}
& \multicolumn{3}{c}{\textbf{TORE}} \\
\cmidrule(lr){2-4}\cmidrule(lr){5-7}\cmidrule(lr){8-10}\cmidrule(lr){11-13}\cmidrule(lr){14-16}
\textbf{Experiment} & \textbf{MSE} & \textbf{EMD} & \textbf{SSIM} & \textbf{MSE} & \textbf{EMD} & \textbf{SSIM} & \textbf{MSE} & \textbf{EMD} & \textbf{SSIM} & \textbf{MSE} & \textbf{EMD} & \textbf{SSIM} & \textbf{MSE} & \textbf{EMD} & \textbf{SSIM} \\
\midrule
\expone  & 0.0375 & 1.722 & 0.832 & 0.0001 & 1.102 & 0.987 & 0.0646 & 1.763 & 0.747 & 0.0080 & 1.721 & 0.977 & 0.2302 & \textendash & 0.623 \\
\exptwo  & 0.0039 & 1.238 & 0.903 & 0.0001 & 0.288 & 0.986 & 0.0055 & 1.280 & 0.883 & 0.0009 & 1.235 & 0.983 & 0.0326 & \textendash & 0.848 \\
\expthree& 0.0236 & 4.569 & 0.901 & 0.0001 & 0.642 & 0.990 & 0.0435 & 4.461 & 0.848 & 0.0050 & 4.571 & 0.986 & 0.1333 & \textendash & 0.762 \\
\expfour & 0.0191 & 2.050 & 0.737 & 0.0004 & 0.409 & 0.948 & 0.0303 & 2.125 & 0.643 & 0.0042 & 2.108 & 0.964 & 0.1413 & \textendash & 0.489 \\
\expfive & 0.0216 & 2.448 & 0.858 & 0.0003 & 0.852 & 0.967 & 0.0367 & 2.449 & 0.782 & 0.0046 & 2.452 & 0.981 & 0.1393 & \textendash & 0.657 \\
\expsix  & 0.0093 & 0.857 & 0.841 & 0.0010 & 0.814 & 0.958 & 0.0171 & 0.870 & 0.795 & 0.0013 & 0.862 & 0.960 & 0.0597 & \textendash & 0.733 \\
\expseven& 0.0080 & 3.577 & 0.858 & 0.0012 & 1.359 & 0.936 & 0.0147 & 3.613 & 0.830 & 0.0011 & 3.561 & 0.966 & 0.0503 & \textendash & 0.769 \\
\expeight& 0.0041 & 1.097 & 0.931 & 0.0011 & 1.078 & 0.967 & 0.0070 & 1.063 & 0.921 & 0.0006 & 1.118 & 0.983 & 0.0251 & \textendash & 0.887 \\
\expnine & 0.0041 & 4.864 & 0.913 & 0.0011 & 0.826 & 0.953 & 0.0080 & 4.965 & 0.894 & 0.0006 & 4.958 & 0.978 & 0.0269 & \textendash & 0.859 \\
\bottomrule
\end{tabular}
\end{adjustbox}
\end{table*}

\begin{table}[!t]
\centering
\small 
\caption{Performance comparison of event-driven encoders (M=8)}\vspace{-0.25em}
\begin{tabular}{lccc}
\hline
Method & Time (ms) & Speed (kevents/s) & Relative Efficiency \\
\hline
DCT & 1.5$\pm$0.5 & 2157.3$\pm$559.5 & 100.0\% \\
DTFT & 3.8$\pm$0.3 & 802.8$\pm$59.0 & 37.2\% \\
DWT & 3.8$\pm$0.2 & 794.4$\pm$40.0 & 36.8\% \\
\hline
\end{tabular}
\label{tab:m8_performance}
\end{table}

\begin{table}[!tp]
\caption{Scene-wise comparison of compression techniques (M=16). IoU is measured on event segmentation using EventSAM.}
\centering
\small 
\setlength{\tabcolsep}{3pt} 
\renewcommand{\arraystretch}{1.1} 
\begin{tabular}{lccccc}
\hline
\textbf{Sequence} & \textbf{CES} & \textbf{DCT} & \textbf{DTFT} & \textbf{DWT} & \textbf{Voxel} \\
\hline
EHPT-XC Indoor HDR        & 0.732 & 0.734 & 0.732 & 0.714 & \textbf{0.837} \\
EHPT-XC In-Grain          & 0.887 & 0.884 & 0.887 & 0.786 & \textbf{0.929} \\
EHPT-XC HDR Blur          & \textbf{0.867} & 0.863 & \textbf{0.867} & 0.812 & 0.855 \\
EHPT-XC Outdoor           & 0.742 & 0.730 & 0.742 & 0.701 & \textbf{0.823} \\
EHPT-XC Outdoor Low       & 0.689 & 0.672 & 0.689 & 0.664 & \textbf{0.897} \\
\hdashline
MVSEC Indoor      & 0.892 & \textbf{0.937} & 0.892 & 0.558 & 0.283 \\
MVSEC Motorcycle  & 0.672 & \textbf{0.721} & 0.672 & 0.296 & 0.381 \\
MVSEC Day1        & 0.986 & \textbf{0.992} & 0.986 & 0.939 & 0.743 \\
MVSEC Night       & 0.834 & \textbf{0.839} & 0.834 & 0.575 & 0.357 \\
\hline
\end{tabular}
\label{tab:scene_wise_iou1}
\end{table}

\begin{figure*}[!b]
  \centering
  \includegraphics[width=0.98\textwidth]{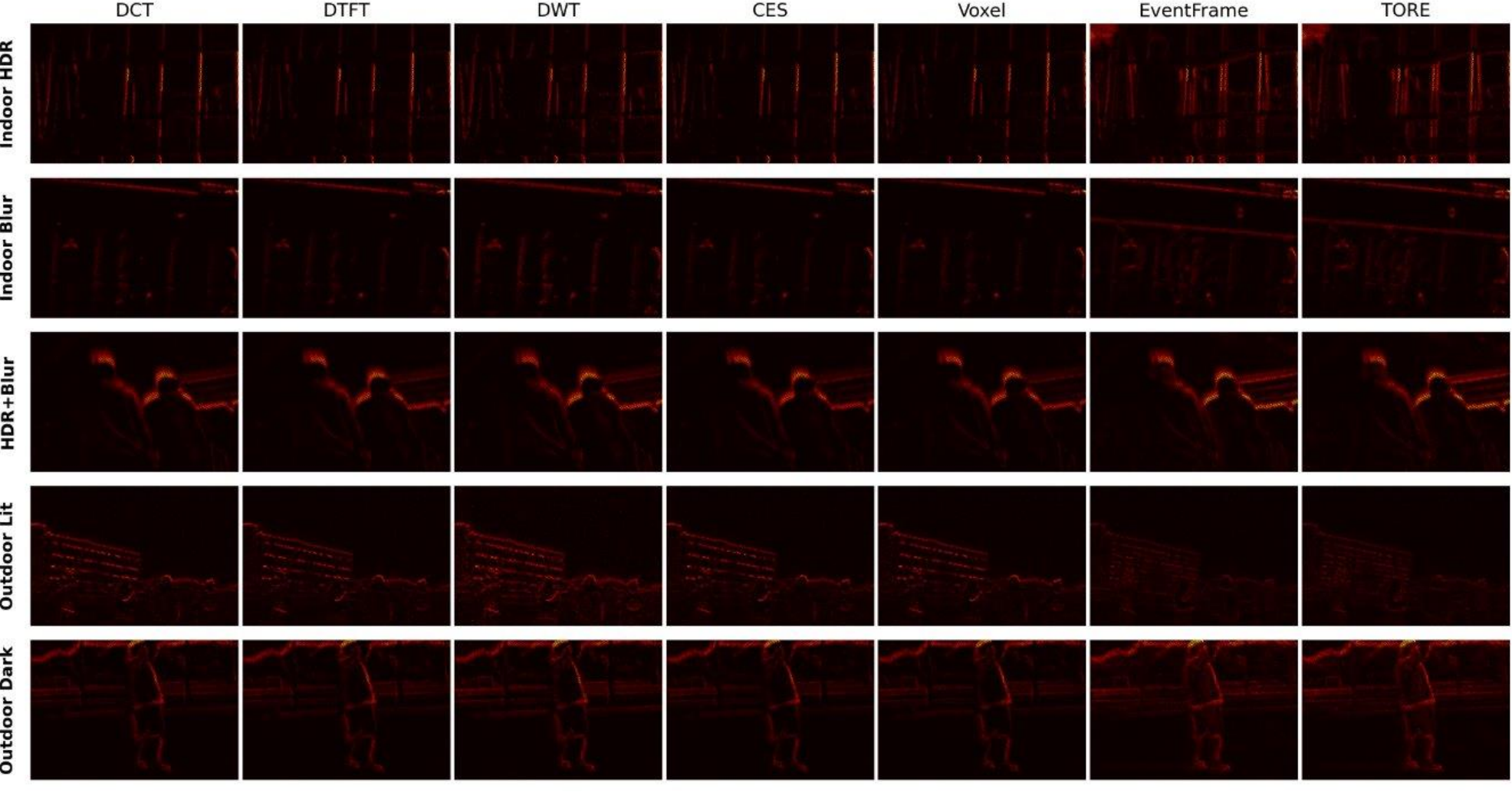}
  \caption{Qualitative results on the EHPT-XC dataset, showing robustness in high-density, real-world scenarios.}
  \label{fig:qualitative_ehpt}
\end{figure*}

\subsection{Adaptive Coefficient Management}

Table~\ref{tab:overall-means-M} reveals how coefficient budget size affects reconstruction quality across transform types. EMD is essentially unchanged across \(M\) for all transforms \((\approx 0.964)\), indicating that coefficient size primarily affects pixel-space fidelity rather than temporal alignment preservation. This finding validates our approach of using fixed temporal windows with variable coefficient budgets.

EECVS's DTFT component benefits significantly from larger coefficient sets. Increasing from $M=8$ to $M=24$ reduces MSE by 2.8\% while simultaneously improving SSIM from 0.882 to 0.887. DCT demonstrates remarkable stability across budgets variations, with MSE varying minimally $(1.731 \text{ vs } 1.740 \times 10^{-3})$. This enables EECVS to deploy DCT efficiently when density demands it. DWT exhibits negligible sensitivity to coefficient count, allowing EECVS to use compact representations for sparse windows without quality degradation.

\begin{table}[ht]
\caption{Scene-wise comparison of compression techniques (M=24). IoU is measured on event segmentation using EventSAM.}
\centering
\small 
\setlength{\tabcolsep}{5pt} 
\renewcommand{\arraystretch}{1.05} 
\begin{tabular}{lccccc}
\hline
\textbf{Sequence} & \textbf{CES} & \textbf{DCT} & \textbf{DTFT} & \textbf{DWT} & \textbf{Voxel} \\
\hline
EHPT-XC Indoor HDR        & 0.739 & 0.733 & 0.739 & 0.706 & \textbf{0.766} \\
EHPT-XC In-Grain          & 0.884 & 0.884 & 0.884 & 0.785 & \textbf{0.934} \\
EHPT-XC HDR Blur          & \textbf{0.867} & 0.863 & \textbf{0.867} & 0.809 & 0.858 \\
EHPT-XC Outdoor           & 0.733 & 0.730 & 0.733 & 0.689 & \textbf{0.855} \\
EHPT-XC Outdoor Low       & 0.688 & 0.678 & 0.688 & 0.658 & \textbf{0.839} \\
\hdashline
MVSEC Indoor      & \textbf{0.932} & \textbf{0.932} & 0.925 & 0.560 & 0.381 \\
MVSEC Motorcycle  & \textbf{0.717} & 0.710 & \textbf{0.717} & 0.294 & 0.300 \\
MVSEC Day1        & \textbf{0.987} & \textbf{0.987} & \textbf{0.987} & 0.940 & 0.719 \\
MVSEC Night       & \textbf{0.842} & 0.840 & \textbf{0.842} & 0.573 & 0.364 \\
\hline
\end{tabular}
\label{tab:scene_wise_iou2}
\end{table}

\begin{figure*}[!h]
  \centering
  \includegraphics[width=0.99\textwidth]{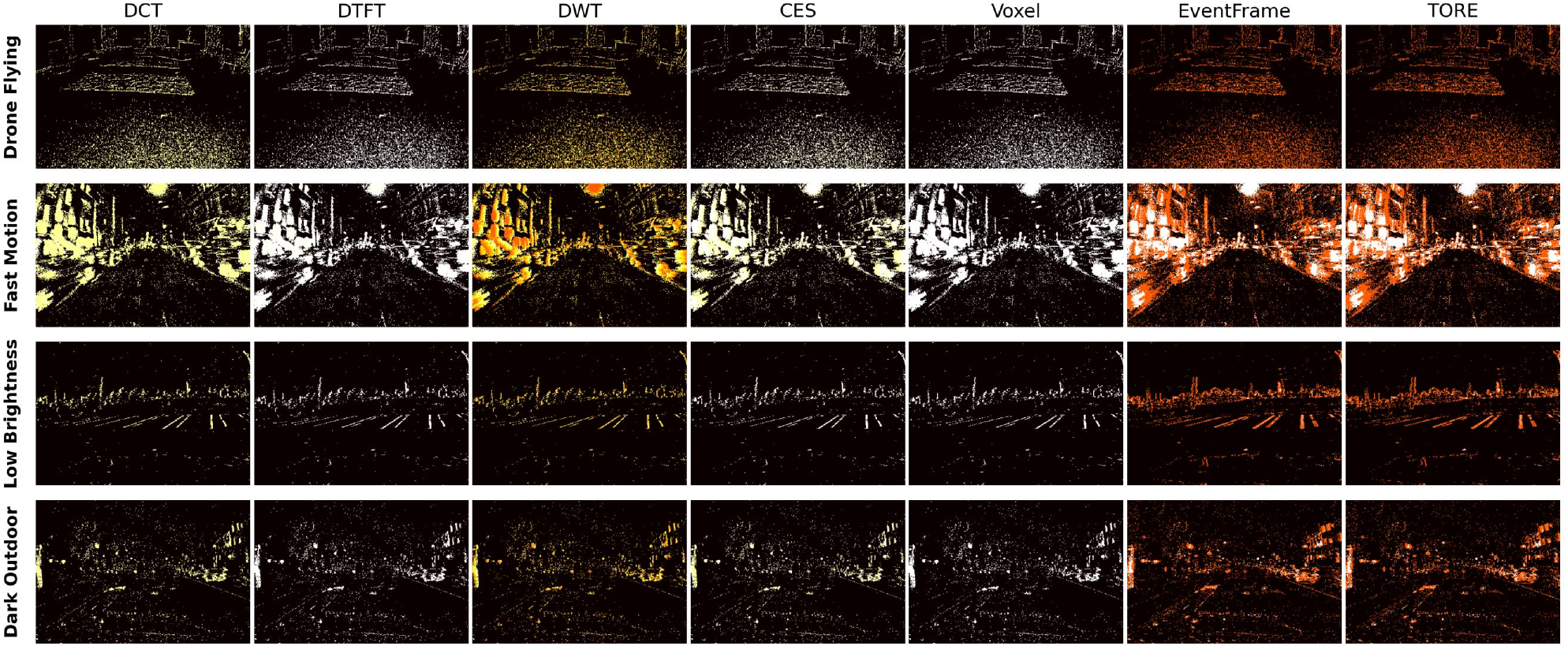}
  \caption{Qualitative results on the MVSEC dataset, demonstrating preservation of temporal fidelity in dynamic scenes.}
  \label{fig:qualitative_mvsec}
\end{figure*}

\subsection{Computational Performance}
Table~\ref{tab:m8_performance} shows significant computational performance differences across transform types with small coefficient budgets (\(M{=}8\)). the DCT encoder achieves the lowest latency (\(1.5\pm0.5\) ms) and the highest throughput (\(2157.3\pm559.5\) kev/s), making it about \(2.7\times\) faster than both DTFT (\(\sim 803\) kev/s) and DWT (\(\sim 794\) kev/s). The relative efficiency column reflects this gap, with DTFT and DWT operating at roughly \(37\%\) of the DCT throughput. While DTFT and DWT exhibit narrower variance in time and speed, DCT still retains substantial headroom even at one standard deviation below its mean performance. DCT still sustains \(\gtrsim\!1.6\) Mev/s. These measurements validate our scheduling policy: routing high-density or latency-constrained windows to DCT processing while reserving DTFT and DWT for scenarios where their reconstruction characteristics justify additional computational cost.

\subsection{Downstream Task Validation}

\textbf{Downstream evaluation protocol.}
To isolate the impact of compression on segmentation, we use a compression--decompression pipeline. We first compress the input event stream with each method, then reconstruct (\emph{decompress}) an event stream by applying the corresponding inverse transform using the retained coefficients. The reconstructed events are then passed to EventSAM~\cite{chen_segment_2024} using the same preprocessing and inference settings across all methods.

EventSAM is not retrained or fine-tuned for any representation. We use the original pretrained weights from~\cite{chen_segment_2024}. Therefore, performance differences reflect the effect of compression and reconstruction rather than task-specific adaptation.

On EHPT scenes, voxelization achieves the highest per-scene IoU. However, EECVS remains highly competitive. The framework's DCT selection achieves mean IoU 0.78 at 24 channels. This falls within seven points of voxels' 0.85 while providing significantly superior computational efficiency. EECVS matches CES and DTFT performance on EHPT sequences, demonstrating consistent quality preservation.

The framework reveals superior generalization capabilities on MVSEC datasets. While voxelization performance drops catastrophically across all MVSEC scenes, EECVS maintains robust performance through adaptive transform selection. At 24 channels, the framework achieves mean IoU 0.87 compared to voxelization's 0.44 on identical data. This pattern persists across channel configurations, confirming EECVS's adaptive approach provides more generalizable representations than fixed methods.

EECVS demonstrates remarkable consistency across channel configurations. The framework's overall mean IoU remains stable (0.82 at both 16 and 24 channels), while voxelization exhibits dataset sensitivity regardless of channel count. These results confirm that EECVS provides balanced, generalizable representations through intelligent adaptive selection.

\subsection{Qualitative Assessment}

Figures~\ref{fig:qualitative_mvsec}-~\ref{fig:qualitative_ehpt} demonstrate EECVS's reconstruction quality across diverse scenarios. The framework's outputs maintain visual similarity to original data at standard viewing distances, with differences concentrated around high-contrast edges and rapid motion regions. DTFT selection preserves edge continuity and phase alignment effectively, while DWT mode maintains isolated contours in sparse windows. DCT occasionally introduces minimal ringing artifacts but provides superior computational efficiency. Overall visual distinctions remain modest, validating the framework's adaptive compression approach while maintaining perceptual quality.

\section{Conclusion}
We presented EECVS, the first adaptive event compression framework that fundamentally advances event camera integration in robotics through three core technical innovations. Our continuous-time Dirac impulse model eliminates temporal binning artifacts inherent in existing approaches, enabling precise transform evaluation at exact event timestamps. The density-driven adaptive selection mechanism intelligently chooses among DCT, DTFT, and DWT based on real-time event characteristics, while transform-specific coefficient pruning strategies optimize compression for each domain's sparsity properties. Together, these innovations produce dense representations that preserve both temporal precision and spatial detail while maintaining computational efficiency.

Our evaluation demonstrates EECVS's superior adaptive capabilities: DTFT achieved the lowest earth mover distance in eight scenarios while maintaining excellent MSE and SSIM performance. DCT reached 1.5 millisecond latency with approximately 2.7 times the throughput of alternative transforms at eight coefficients. On cross-dataset evaluation, the framework achieved mean IoU 0.87 at 24 channels on MVSEC while voxel representations managed only 0.44, demonstrating superior generalization capabilities. On EHPT-XC, performance remained competitive within seven points of voxels while providing computational advantages.

This work establishes adaptive compression as a critical capability for robust event camera deployment in robotics. By intelligently matching compression strategies to event stream characteristics, EECVS bridges the gap between event cameras' unique sensing capabilities and standard perception pipelines' dense input requirements. Future extensions will incorporate learned compression strategies, adaptive temporal windowing, and batch optimization for enhanced performance across even broader robotic applications.

\bibliographystyle{IEEEtran}
\bibliography{references-2,references,ieee-bstctl}


\end{document}